\documentclass[11pt,a4paper]{article}
\usepackage[hyperref]{acl2021}
\usepackage{times}
\usepackage{latexsym}
\usepackage{graphicx}
\usepackage{multirow}
\usepackage{booktabs}
\usepackage{makecell}
\usepackage{float}
\usepackage{comment}
\usepackage[ruled,vlined]{algorithm2e}
\usepackage{amssymb,amsmath,bbm}

\usepackage{xcolor}

\usepackage{microtype}

\aclfinalcopy 

\title{Deriving Word Vectors from Contextualized Language Models\\ using Topic-Aware Mention Selection}

\author{Yixiao Wang\textsuperscript{1}, Zied Bouraoui\textsuperscript{2}, Luis Espinosa Anke\textsuperscript{1} \and Steven Schockaert\textsuperscript{1}\\
\textsuperscript{1}School of Computer Science \& Informatics, Cardiff University, UK \\
\textsuperscript{2} CRIL-CNRS, Universit\'e d'Artois, France\\
\{wangy306,espinosa-ankel,schockaerts1\}@cardiff.ac.uk, 
zied.bouraoui@cril.fr}

\begin{document}

\maketitle

\begin{abstract}
One of the long-standing challenges in lexical semantics consists in learning representations of words which reflect their semantic properties. The remarkable success of word embeddings for this purpose suggests that high-quality representations can be obtained by summarizing the sentence contexts of word mentions. In this paper, we propose a method for learning word representations that follows this basic strategy, but differs from standard word embeddings in two important ways. First, we take advantage of contextualized language models (CLMs) rather than bags of word vectors to encode contexts. Second, rather than learning a word vector directly, we use a topic model to partition the contexts in which words appear, and then learn different topic-specific vectors for each word. Finally, we use a task-specific supervision signal to make a soft selection of the resulting vectors. We show that this simple strategy leads to high-quality word vectors, which are more predictive of semantic properties than word embeddings and existing CLM-based strategies.
\end{abstract}

\section{Introduction}
In the last few years, contextualized language models (CLMs) such as BERT \cite{DBLP:conf/naacl/DevlinCLT19} have largely replaced the use of static (i.e.\ non-contextualized) word vectors in many Natural Language Processing (NLP) tasks. However, static word vectors remain important in applications where word meaning has to be modelled in the absence of (sentence) context. For instance, static word vectors are needed for zero-shot image classification \cite{socher2013zero} and zero-shot entity typing \cite{ma2016label}, for ontology alignment \cite{kolyvakis2018deepalignment} and completion \cite{li2019ontology}, taxonomy learning \cite{bordea-etal-2015-semeval,bordea2016semeval}, or for representing query terms in information retrieval systems \cite{nikolaev2020joint}. Moreover, \citet{liu2020towards} recently found that static word vectors can complement CLMs, by serving as anchors for contextualized vectors,  while \citet{alghanmi2020combining} found that incorporating static word vectors could improve the performance of BERT for social media classification.

Given the impressive performance of CLMs across many NLP tasks, a natural question is whether such models can be used to learn high-quality static word vectors, and whether the resulting vectors have any advantages compared to those from standard word embedding models  \cite{DBLP:journals/corr/abs-1301-3781,DBLP:conf/emnlp/PenningtonSM14}. A number of recent works have begun to explore this question \cite{DBLP:conf/emnlp/Ethayarajh19,DBLP:conf/acl/BommasaniDC20,DBLP:conf/emnlp/VulicPLGK20}. Broadly speaking, the idea is to construct a static word vector for a word $w$ by randomly selecting sentences in which this word occurs, and then averaging the contextualized representations of $w$ across these sentences. 

Since it is not usually computationally feasible to run the CLM on all sentences mentioning $w$, a sample of such sentences has to be selected. This begs the question: how should these sentences be chosen? In the aforementioned works, sentences are selected at random, but this may not be optimal. If we want to use the resulting word vectors in downstream tasks such as zero-shot learning or ontology completion, we need vectors that capture the salient semantic properties of words. Intuitively, we should thus favor sentences that best reflect these properties. For instance, many of the mentions of the word \emph{banana} on Wikipedia are about the cultivation and export of bananas, and about the specifics of particular banana cultivars. By learning a static word vector from such sentences, we may end up with a vector that does not reflect our commonsense understanding of bananas, e.g.\ the fact that they are curved, yellow and sweet.

The main aim of this paper is to analyze to what extent topic models such as Latent Dirichlet Allocation \cite{DBLP:journals/jmlr/BleiNJ03} can be used to address this issue. Continuing the previous example, we may find that the word \emph{banana} occurs in Wikipedia articles on the following topics: economics, biology, food or popular culture. While most mentions might be in articles on economics and biology, it is the latter two topics that are most relevant for modelling the commonsense properties of bananas. Note that the optimal selection of topics is task-dependent, e.g.\ in an NLP system for analyzing financial news, the economics topic would clearly be more relevant. For this reason, we propose to learn a word vector for each topic separately. Since the optimal choice of topics is task-dependent, we then rely on a task-specific supervision signal to make a soft selection of these topic-specific vectors.

Another important question is how CLMs should be used to obtain contextualized word vectors. Given a sentence mentioning $w$, a model such as BERT-base constructs 12 vector representations of $w$, i.e.\ one for each layer of the transformer stack. Previous work has suggested to use the average of particular subsets of these vectors. In particular, \citet{DBLP:conf/emnlp/VulicPLGK20} found that lexical semantics is most prevalent in the representations from the early layers, and that averaging vectors from the first few layers seems to give good results on many benchmarks. On the other hand, these early layers are least affected by the sentence context \cite{DBLP:conf/emnlp/Ethayarajh19}, hence such strategies might not be suitable for learning topic-specific vectors. We therefore also explore a different strategy, which is to mask the target word in the given sentence, i.e.\ to replace the entire word by a single [MASK] token, and to use the vector representation of this token at the final layer. The resulting vector representations thus specifically encode what the given sentence reveals about the target word, making this a natural strategy for learning topic-specific vectors. 

Note that there is a clear relationship between this latter strategy and CBOW \cite{DBLP:journals/corr/abs-1301-3781}: where in CBOW the vector representation of $w$ is obtained by averaging the vector representations of the context words that co-occur with $w$, we similarly represent words by averaging context representations. The main advantage compared to CBOW thus comes from the higher-quality context encodings that can be obtained using CLMs. The main challenge, as already mentioned, is that we cannot consider all the mentions of $w$, whereas this is typically feasible for CBOW (and other standard word embedding models).
%
Our contributions can be summarized as follows\footnote{All code and data to replicate our experiments is available at \url{https://github.com/Activeyixiao/topic-specific-vector/}.}:

\begin{itemize}
\item We analyze different strategies for deriving word vectors from CLMs,  which rely on sampling mentions of the target word from a text collection.
\item We propose the use of topic models to improve how these mentions are sampled. In particular, rather than learning a single vector representation for the target word, we learn one vector for each sufficiently relevant topic.
\item We propose to construct the final representation of a word $w$ as a weighted average of different vectors. This allows us to combine multiple vectors without increasing the dimensionality of the final representations. We use this approach for combining different topic-specific vectors and for combining vectors from different transformer layers.
\end{itemize}

\section{Related Work}

A few recent works have already proposed strategies for computing static word vectors from CLMs. While \citet{DBLP:conf/emnlp/Ethayarajh19} relied on principal components of individual transformer layers for this purpose, most approaches rely on averaging the contextualised representations of randomly selected mentions of the target word \cite{DBLP:conf/acl/BommasaniDC20,DBLP:conf/emnlp/VulicPLGK20}.  Several authors have pointed out that the representations obtained from early layers tend to perform better in lexical semantics probing tasks. However, \citet{DBLP:conf/acl/BommasaniDC20} found that the optimal layer depends on the number of sampled mentions, with later layers performing better when a large number of mentions is used. Rather than fixing a single layer, \citet{DBLP:conf/emnlp/VulicPLGK20} advocated averaging representations from several layers. Note that none of the aforementioned methods uses masking when computing contextualized vectors. This means that the final representations may have to be obtained by pooling different word-piece vectors, usually by averaging them.

As an alternative to using topic models, \citet{chronis-erk-2020-bishop} cluster the contextual word vectors, obtained from mentions of the same word. The resulting multi-prototype representation is then used to compute word similarity in an adaptive way. Along similar lines, \citet{amrami2019towards} cluster contextual word vectors for word sense induction. \citet{DBLP:journals/corr/abs-2010-12626} showed that clustering the contextual representations of a given set of words can produce clusters of semantically related words, which were found to be similar in spirit to LDA topics.
The idea of learning topic-specific representations of words has  been extensively studied in the context of standard word embeddings \cite{DBLP:conf/aaai/LiuLCS15,DBLP:conf/acl/LiCZM16,DBLP:conf/sigir/ShiLJSL17,DBLP:journals/tacl/ZhuZH20}. To the best of our knowledge, learning topic-specific word representations using CLMs has not yet been studied. More broadly, however, some recent methods have combined CLMs with topic models. For instance, \citet{DBLP:conf/acl/PeineltNL20} use such a combination for predicting semantic similarity. In particular they use the LDA or GSDMM topic distribution of two sentences to supplement their BERT encoding. Finally, \citet{DBLP:journals/corr/abs-2004-03974} suggested using sentence embeddings from SBERT \cite{reimers-gurevych-2019-sentence} as input to a neural topic model, with the aim of learning more coherent topics.

\section{Constructing Word Vectors}
In Section \ref{secObtainingContextualiseVectors}, we first describe different strategies for deriving static word vectors from CLMs. Section \ref{secSelectingTopicMentions} subsequently describes how we choose the most relevant topics for each word, and how we sample topic-specific word mentions. Finally, in Section \ref{secAveragingModel} we explain how the resulting topic-specific representations are combined to obtain task-specific word vectors.
\subsection{Obtaining Contextualized Word Vectors}\label{secObtainingContextualiseVectors}
We first briefly recall the basics of the BERT contextualised language model. BERT represents a sentence $s$ as a sequence of word-pieces $w_1...,w_n$. Frequent words will typically be represented as a single word-piece, but in general, word-pieces may correspond to sub-word tokens. Each of these word-pieces $w$ is represented as an input vector, which is constructed from a static word-piece embedding $\mathbf{w}_0$ (together with vectors that encode at which position in the sentence the word appears, and in which sentence). The resulting sequence of word-piece vectors is then fed to a stack of 12 (for BERT-base) or 24 (for BERT-large) transformer layers. Let us write $\mathbf{w}_i^s$ for the representation of word-piece $w$ in the i\textsuperscript{th} transformer layer. We will refer to the representation in the last layer, i.e.\ $\mathbf{w}_{12}^s$ for BERT-base and $\mathbf{w}_{24}^s$ for BERT-large, as the output vector. When BERT is trained, some of the word-pieces are replaced by a special [MASK] token. The corresponding output vector then encodes a prediction of the masked word-piece.

Given a sentence $s$ in which the word $w$ is mentioned, there are several ways in which BERT and related models can be used to obtain a vector representation of $w$. If $w$ consists of a single word-piece, a natural strategy is to feed the sentence $s$ as input and use the output vector as the representation of $w$. However, several authors have found that it can be beneficial to also take into account some or all of the earlier transformer layers, where fine-grained word senses are mostly captured in the later layers \cite{DBLP:conf/nips/ReifYWVCPK19} but word-level lexical semantic features are primarily found in the earlier layers \cite{DBLP:conf/emnlp/VulicPLGK20}. For this reason, we will also experiment with models in which the vectors $\mathbf{w}_{1}^s,...,\mathbf{w}_{12}^s$ (or $\mathbf{w}_{1}^s,...,\mathbf{w}_{24}^s$ in the case of BERT-large) are all used. In particular, our model will construct a weighted average of these vectors, where the weights will be learned from training data (see Section \ref{secAveragingModel}). For words that consist of multiple word-pieces, following common practice, we compute the representation of $w$ as the average of its word-piece vectors. For instance, this strategy was found to outperform other aggregation strategies in \citet{DBLP:conf/acl/BommasaniDC20}. 

We will also experiment with a strategy that relies on masking. In this case, the word $w$ is replaced by a single [MASK] token (even if $w$ would normally be tokenized into more than one word-piece). Let us write $\mathbf{m}_w^s$ for the output vector corresponding to this [MASK] token. Since this vector corresponds to BERT's prediction of what word is missing, this vector should intuitively capture the properties of $w$ that are asserted in the given sentence. We can thus expect that these vectors $\mathbf{m}_w^s$ will be more sensitive to how the sentences mentioning $w$ are chosen. Note that in this case, we only use the output layer, as the earlier layers are less likely to be informative. 

To obtain a static representation of $w$, we first select a set of sentences $s_1,...,s_n$ in which $w$ is mentioned. Then we compute vector representations $\mathbf{w}^{s_1},...,\mathbf{w}^{s_n}$ of $w$ from each of these sentences, using any of the aforementioned strategies. Our final representation $\mathbf{w}$ is then obtained by averaging these sentence-specific representations, i.e.:
$$
\mathbf{w} = \frac{\sum_{i=1}^n \mathbf{w}^{s_i}}{\|\sum_{i=1}^n \mathbf{w}^{s_i}\|}
$$

\subsection{Selecting Topic-Specific Mentions}\label{secSelectingTopicMentions}
To construct a vector representation of $\mathbf{w}$, we need to select some sentences $s_1,...,s_n$ mentioning $w$. While these sentences are normally selected randomly, our hypothesis in this paper is that purely random strategies may not be optimal. Intuitively, this is because the contexts in which a given word $w$ is most frequently mentioned might not be the most informative ones, i.e.\ they may not be the contexts which best characterize the properties of $w$ that matter for a given task.
To test this hypothesis, we experiment with a strategy based on topic models. Our strategy relies on the following steps:
\begin{enumerate}
\item Identify the topics which are most relevant for the target word $w$;
\item For each of the selected topics $t$, select sentences $s^t_1,...,s^t_n$ mentioning $w$ from documents that are closely related to this topic.
\end{enumerate}
For each of the selected topics $t$, we can then use the sentences $s^t_1,...,s^t_n$ to construct a topic-specific vector $\mathbf{w}^t$, using any of the strategies from Section \ref{secObtainingContextualiseVectors}. The final representation of $w$ will be computed as a weighted average of these topic-specific vectors, as will be explained in Section \ref{secAveragingModel}. 

We now explain these two steps in more detail. First, we use Latent Dirichlet Allocation (LDA)  \cite{DBLP:journals/jmlr/BleiNJ03} to obtain a representation of each document $d$ in the considered corpus as a multinomial distribution over $m$ topics. Let us write $\tau_i(d)$ for the weight of topic $i$ in the representation of document $d$, where $\sum_{i=1}^m \tau_i(d)=1$. Suppose that the word $w$ is mentioned $N_w$ times in the corpus, and let $d^w_j$ be the document in which the $j$\textsuperscript{th} mention of $w$ occurs. Then we define the importance of topic $i$ for word $w$ as follows:
\begin{align}\label{eqTopicWordImportanceDef}
\tau_i(w) = \frac{1}{N_w} \sum_{j=1}^{N_w} \tau_i(d_j^w)
\end{align}
In other words, the importance of topic $i$ for word $w$ is defined as the average importance of topic $i$ for the documents in which $w$ occurs. To select the set of topics $\mathcal{T}_w$ that are relevant to $w$, we rank the topics from most to least important and then select the smallest set of topics whose cumulative importance is at least ${60}\%$, i.e.\ $\mathcal{T}_w$ is the smallest set of topics such that $\sum_{t_i\in \mathcal{T}_w}\tau_i(w) \geq 0.6$.

For each of the topics $t_i$ in $\mathcal{T}_w$ we select the corresponding sentences $s^t_1,...,s^t_n$ as follows. We rank all the documents in which $w$ is mentioned according to $\tau_i(d)$. Then, starting with the document with the highest score (i.e.\ the document for which topic $i$ is most important), we iterate over the ranked list of documents, selecting all sentences from these documents in which $w$ is mentioned, until we have obtained a total of $n$ sentences.

\subsection{Combining Word Representations} \label{secAveragingModel}

Section \ref{secObtainingContextualiseVectors} highlighted a number of strategies that could be used to construct a vector representation of a target word $w$. As mentioned before, it can be beneficial to combine vector representations from different transformer layers. To this end, we propose to learn a weighted average of the different input vectors, using a task specific supervision signal. In particular, let $\mathbf{w}_1,...,\mathbf{w}_k$ be the different vector representations we have available for word $w$ (e.g.\ the vectors from different transformer layers). To combine these vectors, we compute a weighted average as follows:
\begin{align}
\lambda_i &= \frac{\exp(a_i)}{\sum_{j=1}^k \exp(a_i)}\label{eqSingleAttention1}\\
\mathbf{w} &= \frac{\sum_i \lambda_i \mathbf{w}_i}{\|\sum_i \lambda_i \mathbf{w}_i\|} \label{eqSingleAttention2}
\end{align}
where the scalar parameters $a_1,...a_k\in\mathbb{R}$ are jointly learned with the model in which $\mathbf{w}$ is used. Another possibility would be to concatenate the input vectors $\mathbf{w}_1,...,\mathbf{w}_k$. However, this significantly increases the dimensionality of the word representations, which can be challenging in downstream applications. In initial experiments, we also confirmed that this concatenation strategy indeed under-performs the use of weighted averages.

If topic-specific vectors are used, we also want to compute a weighted average of the available vectors. However, \eqref{eqSingleAttention1}--\eqref{eqSingleAttention2} cannot be used in this case, because the set of topics for which topic-specific vectors are available differs from word to word.
 Let us write $\mathbf{w}^i_{\textit{topic}}$ for the representation of word $w$ that was obtained for topic $t_i$, where we assume $\mathbf{w}^i_{\textit{topic}}=\mathbf{0}$ if $t_i\notin \mathcal{T}_w$. We then define:
\begin{align}
\mu_i^w &= \frac{\exp(b_i)\cdot \mathbbm{1}[t_i\in \mathcal{T}_w]}{\sum_{j=1}^k \exp(b_i)\cdot \mathbbm{1}[t_j\in \mathcal{T}_w]}\label{eqDoubleAttention1}\\
\mathbf{w}_{\textit{topic}} &= \frac{\sum_i \mu_i^w \mathbf{w}^i_{\textit{topic}}}{\| \sum_i \mu_i^w \mathbf{w}^i_{\textit{topic}}\|}\label{eqDoubleAttention2}
\end{align}
where $\mathbbm{1}[t_i\in \mathcal{T}_w]=1$ if topic $t_i$ is considered to be relevant for word $w$ (i.e.\ $t_i\in\mathcal{T}_w$), and $\mathbbm{1}[t_i\in \mathcal{T}_w]=0$ otherwise. Note that the softmax function in \eqref{eqDoubleAttention1} relies on the scalar parameters $b_1,...,b_k\in \mathbb{R}$, which are independent of $w$. However, the softmax is selectively applied to those topics that are relevant to $w$, which is why the resulting weight $\mu_i^w$ is dependent on $w$, or more precisely, on the set of topics $\mathcal{T}_w$.

\section{Evaluation}
We compare the proposed strategy with standard word embeddings and existing CLM-based strategies. In Section \ref{secSetup} we first describe our experimental setup. Section \ref{secDatasets} then provides an overview of the datasets we used for the experiments, where we focus on lexical classification benchmarks. These benchmarks in particular allow us to assess how well various semantic properties can be predicted from the word vectors. The experimental results are discussed in Section \ref{secResults} and a qualitative analysis is presented in Section \ref{sec:qualitative}.

\subsection{Experimental Setup}\label{secSetup}
We experiment with a number of different strategies for obtaining word vectors:
\begin{description}
\item[\textbf{C\textsubscript{last}}] We take the vector representation of $w$ from the last transformer layer (i.e.\ $\mathbf{w}_{12}^s$ or $\mathbf{w}_{24}^s$). 
\item[\textbf{C\textsubscript{input}}] We take the input embedding of $w$ (i.e.\ $\mathbf{w}_{0}$).
\item[\textbf{C\textsubscript{avg}}] We take the average of $\mathbf{w}_{0},\mathbf{w}_{1}^s,...,\mathbf{w}_{12}^s$ for the \emph{base} models and $\mathbf{w}_{0},\mathbf{w}_{1}^s,...,\mathbf{w}_{24}^s$ for the \emph{large} models.
\item[\textbf{C\textsubscript{all}}] We use all of $\mathbf{w}_{0},\mathbf{w}_{1}^s,...,\mathbf{w}_{12}^s$ as input for the \emph{base} models, and all of $\mathbf{w}_{0},\mathbf{w}_{1}^s,...,\mathbf{w}_{24}^s$ for the \emph{large} models. These vectors are then aggregated using \eqref{eqSingleAttention1}--\eqref{eqSingleAttention2}, i.e.\ we use a learned soft selection of the transformer layers.
\item[\textbf{C\textsubscript{mask}}] We replace the target word by [MASK] and use the corresponding output vector.
\end{description}
For words consisting of more than one word-piece, we average the corresponding vectors in all cases, except for \textbf{C\textsubscript{mask}} where we always end up with a single vector (i.e.\ we replace the entire word by a single [MASK] token). 
We also consider three variants that rely on topic-specific vectors:
\begin{description}
\item[\textbf{T\textsubscript{last}}] We learn topic-specific vectors using the last transformer layers. These vectors are then used as input to \eqref{eqDoubleAttention1}--\eqref{eqDoubleAttention2}.
\item[\textbf{T\textsubscript{avg}}] Similar to the previous case but using the average of all transformer layers.
\item[\textbf{T\textsubscript{mask}}] Similar to the previous cases but using the output vector of the masked word mention.
\end{description}
Furthermore, we consider variants of $\textbf{T\textsubscript{last}}$, $\textbf{T\textsubscript{avg}}$ and $\textbf{T\textsubscript{mask}}$ in which a standard (i.e.\ unweighted) average of the available topic-specific vectors is computed, instead of relying on \eqref{eqDoubleAttention1}--\eqref{eqDoubleAttention2}. We will refer to these averaging-based variants as $\textbf{A\textsubscript{last}}$, $\textbf{A\textsubscript{avg}}$ and $\textbf{A\textsubscript{mask}}$.
As baselines, we also consider the two Word2vec models \cite{DBLP:journals/corr/abs-1301-3781}: 
\begin{description}
\item[SG] 300-dimensional Skip-gram vectors trained on a May 2016 dump of the English Wikipedia, using a window size of 5 tokens, and minimum frequency threshold of 10.
\item[CBOW] 300-dimensional Continuous Bag-of-Words vectors trained on the same corpus and with the same hyperparameters as  \textbf{SG}.
\end{description}
We show results for four pre-trained CLMs \cite{DBLP:conf/naacl/DevlinCLT19,DBLP:journals/corr/abs-1907-11692}: BERT-base-uncased, BERT-large-uncased, RoBERTa-base-uncased, RoBERTa-large-uncased\footnote{We used the  implementations from \url{https://github.com/huggingface/transformers}.}. As the corpus for sampling word mentions, we used the same Wikipedia dump as for training the word embeddings models. For \textbf{C\textsubscript{mask}}, \textbf{C\textsubscript{last}}, \textbf{C\textsubscript{avg}} and \textbf{C\textsubscript{all}} we selected 500 mentions. 
For the topic-specific strategies (\textbf{T\textsubscript{last}}, \textbf{T\textsubscript{avg}} and \textbf{T\textsubscript{mask}}) we selected 100 mentions per topic.
To obtain the topic assignments, we used Latent Dirichlet Allocation \cite{DBLP:journals/jmlr/BleiNJ03} with 25 topics. We set $\alpha=0.0001$ to restrict the total number of topics attributed to a document, and use default values for the other hyper-parameters\footnote{We used the implementation from \url{https://radimrehurek.com/gensim/wiki.html}.}. To select the relevant topics for a given word $w$, we find the smallest set of topics whose cumulative importance score $\tau_i(w)$ is at least 60\%, with a maximum of 6 topics. In the experiments, we restrict the vocabulary to those words with at least 100 occurrences in Wikipedia.

\begin{table*}[t]
\centering
\setlength{\tabcolsep}{3pt}
\footnotesize
\begin{tabular}{l l cccc l cccc l cccc l cccc}
\toprule
 && \multicolumn{4}{c}{\textbf{BERT-base}} && \multicolumn{4}{c}{\textbf{BERT-large}} && \multicolumn{4}{c}{\textbf{RoBERTa-base}} && \multicolumn{4}{c}{\textbf{RoBERTa-large}}\\
 \cmidrule(lr){3-6}
 \cmidrule(lr){8-11}
 \cmidrule(lr){13-16}
 \cmidrule(lr){18-21}
 && \textbf{MC} & \textbf{CS} & \textbf{SS} & \textbf{BD} 
 && \textbf{MC} & \textbf{CS} & \textbf{SS} & \textbf{BD}
 && \textbf{MC} & \textbf{CS} & \textbf{SS} & \textbf{BD}
 && \textbf{MC} & \textbf{CS} & \textbf{SS} & \textbf{BD}\\
\midrule
\textbf{SG} && 59.6 & 54.5 & 55.6 & \textbf{49.1} 
&& 59.6 & 54.5 & 55.6 & \textbf{49.1} 
&& 59.6 & 54.5 & 55.6 & \textbf{49.1} 
&& 59.6 & 54.5 & 55.6 & \textbf{49.1}\\ 
\textbf{CBOW} && 61.1 & 50.6 & 48.4 & 45.0 
&& 61.1 & 50.6 & 48.4 & 45.0 
&& 61.1 & 50.6 & 48.4 & 45.0 
&& 61.1 & 50.6 & 48.4 & 45.0\\ 
\midrule
\textbf{C\textsubscript{mask}} && 54.6 & 44.0 & 48.8 & 38.9 
&& 52.0 & 43.0 & 48.7 & 38.7 
&& 56.0 & 43.4 & 47.1 & 42.1 
&& 55.8 & 42.3 & 47.0 & 38.1\\ 
\textbf{C\textsubscript{last}} && 52.9 & 45.1 & 46.7 & 38.4 
&& 54.3 & 46.2 & 48.2 & 39.6 
&& 56.5 & 42.2 & 46.1 & 37.3 
&& 56.3 & 43.8 & 46.5 & 37.8\\ 
\textbf{C\textsubscript{input}} && 48.9 & 32.2 & 41.1 & 34.8 
&& 53.1 & 33.0 & 39.0 & 34.5 
&& 42.1 & 25.6 & 35.2 & 31.8 
&& 51.3 & 31.4 & 28.6 & 36.0\\ 
\textbf{C\textsubscript{avg}} && 45.9 & 32.8 & 44.1 & 36.4 
&& 50.0 & 37.1 & 42.7 & 36.7 
&& 39.4 & 21.6 & 30.8 & 28.7 
&& 43.7 & 22.9 & 30.1 & 28.2\\ 
\textbf{C\textsubscript{all}} && 45.9 & 31.0 & 41.3 & 35.4 
&& 45.0 & 33.7 & 43.4 & 24.6 
&& 32.8 & 19.0 & 25.9 & 24.7 
&& 37.5 & 21.2 & 30.4 & 28.6\\ 
\midrule
\textbf{T\textsubscript{mask}} && 58.6 & 54.1 & 60.1 & 45.8 
&& 62.8 & 54.6 & \textbf{61.4} & 46.2 
&& 56.4 & 49.4 & 56.7 & 42.1 
&& 59.6 & 50.4 & 57.2 & 42.1\\ 
\textbf{T\textsubscript{last}} && \textbf{63.6} & 51.8 & 59.5 & 47.3 
&& 60.5 & 54.8 & 61.2 & 49.2 
&& 52.8 & 40.1 & 54.6 & 41.2 
&& 60.2 & 48.5 & 59.5 & 45.2\\ 
\textbf{T\textsubscript{avg}} && 61.0 & 52.7 & 59.6 & 42.3 
&& 65.2 & 52.4 & 60.7 & 48.4 
&& 54.2 & 39.9 & 55.9 & 41.5 
&& 59.5 & 46.8 & 60.0 & 45.2\\ 
\midrule
\textbf{A\textsubscript{mask}} && 61.6 & 53.5 & 59.6 & 41.5 
&& 63.0 & 56.4 & 60.6 & 41.5 
&& 61.2 & 55.3 & 59.6 & 40.6 
&& 63.4 & \textbf{57.1} & 61.2 & 42.3\\ 
\textbf{A\textsubscript{last}} && 60.8 & 49.6 & 57.9 & 44.4 
&& 61.4 & 55.5 & 60.3 & 46.7 
&& 50.3 & 36.8 & 56.5 & 39. 7
&& 59.5 & 47.3 & 58.0 & 41.2\\ 
\textbf{A\textsubscript{avg}} && 60.7 & 49.7 & 57.9 & 44.4 
&& 63.9 & 52.0 & 59.4 & 44.0 
&& 55.6 & 40.6 & 56.4 & 39.8 
&& 59.4 & 47.3 & 58.0 & 41.2\\ 
\midrule
\textbf{C\textsubscript{mask}-PCA} && 56.8 & 46.4 & 49.2 & 38.8 
&& 56.6 & 43.5 & 48.4 & 39.2 
&& 58.8 & 51.6 & 50.4 & 39.2 
&& 58.3 & 49.8 & 49.3 & 39.3\\ 
\textbf{T\textsubscript{mask}-PCA} && 63.3 & \textbf{56.2} & \textbf{62.6} & 46.9 
&& \textbf{64.4} & \textbf{57.3} & 60.6 & 48.0 
&& \textbf{61.6} & \textbf{55.8} & \textbf{62.5} & 46.0 
&& \textbf{65.4} & 56.3 & \textbf{64.1} & 46.4\\ 
\bottomrule
\end{tabular}
\caption{Results of lexical feature classification experiments for the extended McRae feature norms (MC), CSLB norms (CS), WordNet Supersenses (SS) and BabelNet domains (BD). Results are reported in terms of F1 ($\%$). \label{tabLexicalFeaturesResults}}
\end{table*}
\subsection{Datasets}\label{secDatasets}
\begin{table}[t]
\footnotesize
\centering
\begin{tabular}{@{}llrr@{}}
\toprule   
\textbf{Dataset} & \textbf{Type} & \textbf{Words} & \textbf{Properties}\\[-.5ex]
\midrule
McRae & Commonsense & 475 & 49 \\
CSLB & Commonsense & 570 & 54  \\
WN supersenses  & Taxonomic   & 24,324   & 24\\
BN domains & Topical &  43,319  & 34  \\
\bottomrule   
\end{tabular}
\caption{Overview of the considered datasets.}
\label{datasetstat}
\end{table}

\noindent For the experiments, we focus on a number of lexical classification tasks, where categories of individual words need to be predicted. In particular, we used two datasets which are focused on commonsense properties (e.g.\ \emph{dangerous}): the extension of the McRae feature norms dataset \cite{mcrae2005semantic} that was introduced by \citet{forbes2019neural}\footnote{\url{https://github.com/mbforbes/physical-commonsense}} and the CSLB Concept Property Norms\footnote{\url{https://cslb.psychol.cam.ac.uk/propnorms}}. We furthermore used the WordNet supersenses dataset\footnote{\url{https://wordnet.princeton.edu/download}}, which groups nouns into broad categories (e.g.\ \emph{human}). Finally, we also used the BabelNet domains dataset\footnote{\url{http://lcl.uniroma1.it/babeldomains/}} \cite{camacho2017babeldomains}, which assigns lexical entities to thematic domains (e.g.\ \emph{music}). 

In our experiments, we have only considered properties/classes for which sufficient positive examples are available, i.e.\ at least 10 for McRae, 30 for CSLB, and 100 for WordNet supersenses and BabelNet domains. For the McRae dataset, we used the standard training-validation-test split. For the other datasets, we used random splits of 60\% for training, 20\% for tuning and 20\% for testing. An overview of the datasets is shown in Table \ref{datasetstat}.

For all datasets, we consider a separate binary classification problem for each property and we report the (unweighted) average of the F1 scores for the different properties. To classify words, we feed their word vector directly to a sigmoid classification layer. We optimise the network using AdamW with a cross-entropy loss. The batch size and learning rate were tuned, with possible values chosen from {4,8,16} and {0.01, 0.005, 0.001, 0.0001} respectively. Note that for \textbf{C\textsubscript{all}} and the topic-specific variants, the classification network jointly learns the parameters of the classification layer and the attention weights in \eqref{eqSingleAttention1} and \eqref{eqDoubleAttention1} for combining the input vectors.

\subsection{Results}\label{secResults}
The results are shown in Table \ref{tabLexicalFeaturesResults}. 
We consistently see that the topic-specific variants outperform the different \textbf{C}-variants, often by a substantial margin. This confirms our main hypothesis, namely that using topic models to determine how context sentences are selected has a material effect on the quality of the resulting word representations.
Among the \textbf{C}-variants, the best results are obtained by \textbf{C\textsubscript{mask}} and \textbf{C\textsubscript{last}}. None of the three \textbf{T}-variants consistently outperforms the others. 
Surprisingly, the \textbf{A}-variants outperform the corresponding \textbf{T}-variants in several cases. This suggests that the outperformance of the topic-specific vectors primarily comes from the fact that the context sentences for each word were sampled in a more balanced way (i.e.\ from documents covering a broader range of topics), rather than from the ability to adapt the topic weights based on the task. This is a clear benefit for applications, as the \textbf{A}-variants allow us to simply represent each word as a static word vector.

The performance of SG and CBOW is also surprisingly strong. In particular, these traditional word embedding models outperform all of the \textbf{C}-variants, as well as the \textbf{T} and \textbf{A} variants in some cases, especially for BERT-base and RoBERTa-base. This seems to be related, at least in part, to the lower dimensionality of these vectors. The classification network has to be learned from a rather small number of examples, especially for McRae and CSLB. Having 768 or 1024 dimensional input vectors can be problematic in such cases. To analyse this effect, we used Principal Component Analysis (PCA) to reduce the dimensionality of the CLM-derived vectors to 300. For this experiment, we focused in particular on \textbf{C\textsubscript{mask}} and \textbf{T\textsubscript{mask}}. The results are also shown in Table \ref{tabLexicalFeaturesResults} as \textbf{C\textsubscript{mask}-PCA} and \textbf{T\textsubscript{mask}-PCA}. As can be seen, this dimensionality reduction step has a clearly beneficial effect, with \textbf{T\textsubscript{mask}-PCA} outperforming all baselines, except for the BabelNet domains benchmark. The latter benchmark is focused on thematic similarity rather than semantic properties, which the CLM-based representations seem to struggle with.

\begin{table*}
\footnotesize
\renewcommand{\arraystretch}{1.1}
\centering
\begin{tabular}{lll@{}}
\toprule
\multicolumn{1}{l}{\textbf{\textsc{{word}}}} & \multicolumn{1}{c}{\textbf{\textsc{topic}}}                              & \multicolumn{1}{c}{\textbf{\textsc{nearest neighbours}}}                  \\ \midrule
\multirow{3}{*}{\textbf{partner}}        & \{research, professor, science, education, institute\}   & beneficiary, creditor, investor, employer, stockholder  \\
                                & \{football, republican, coach, senate, representatives\} & lobbyist, bookkeeper, cashier, stockbroker, clerk       \\
                                & \{game, book, novel, story, reception\}                       & nanny, spouse, lover, friend, secretary                 \\ \midrule
\multirow{3}{*}{\textbf{cell}}           & \{protein, disease, medical, cancer, cells\}             & lymphocyte, macrophage, axon, astrocyte, organelle      \\
                                & \{food, plant, water, gas, power, oil\}                  & electrode, electrolyte, cathode, anode, substrate       \\
                                & \{physics, mathematics, space, ngc, theory\}             & surface, torus, mesh, grid, cone                        \\ \midrule
\multirow{3}{*}{\textbf{port}}           & \{station, building, railway, historic, church\}         & harbor, seaport, dock, waterfront, city                 \\
                                & \{radio, station, fm, software, data, forewings\}        & link, gateway, router, line, socket                     \\
                                & \{game, book, novel, story, reception\}            & version, remake, compilation, patch, modification       \\ \midrule
\multirow{3}{*}{\textbf{bulb}}           & \{station, building, railway, historic, church\}         & lamp, transformer, dynamo, projector, lighting          \\
                                & \{protein, disease, medical, cancer, cells\}             & epithelium, ganglion, nucleus, gland, cortex            \\
                                & \{species, genus, described, description, flowers\}      & rootstock, fern, vine, tuber, clover                    \\ \midrule
\multirow{3}{*}{\textbf{mail}}           & \{station, building, railway, historic, church\}         & cargo, grain, baggage, coal, livestock                  \\
                                & \{game, book, novel, story, reception\}            & paper, jewelry, telephone, telegraph, typewriter        \\
                                & \{party, election, minister, elected, elections\}        & telemarketing, spam, wiretap, internet, money           \\ \midrule
\multirow{3}{*}{\textbf{fingerprint}}    & \{radio, station, fm, software, data, forewings\}        & signature, checksum, bitmap, texture, text              \\
                                & \{game, book, novel, story, reception\}            & cadaver, skull, wiretap, body, tooth                    \\
                                & \{party, election, minister, elected, elections\}        & wiretap, forensics, postmortem, polygraph, check        \\ \Xhline{2\arrayrulewidth}
\multirow{3}{*}{\textbf{sky}}            & \{greek, ancient, castle, king, roman\}                  & underworld, sun, afterlife, zodiac, moon                \\
                                & \{river, lake, mountain, island, village\}               & horizon, ocean, earth, sun, globe                       \\
                                & \{physics, mathematics, space, ngc, theory\}             & ionosphere, sun, globe, earth, heliosphere              \\ \midrule
\multirow{3}{*}{\textbf{strength}}       & \{food, plant, water, gas, power\}                       & stiffness, ductility, hardness, permeability, viscosity \\
                                & \{game, book, novel, story, reception\}            & intelligence, agility, charisma, power, telepathy       \\
                                & \{army, regiment, navy, ship, air\}                      & morale, firepower, resistance, force, garrison          \\ \midrule
\multirow{2}{*}{\textbf{noon}}            & \{physics, mathematics, space, ngc, theory\}                  & declination, night, equinox, perihelion, latitude                \\
                                & \{army, regiment, navy, ship, air\}               & dawn, sunset, night, morning, shore                       \\ \midrule
\multirow{2}{*}{\textbf{galaxy}}            & \{physics, mathematics, space, ngc, theory\}                  & nebula, quasar, pulsar, nova, star

                \\
                                & \{game, book, novel, story, reception\}               & globe, future, world, planet, nation
                      \\ \bottomrule

\end{tabular}
\caption{Nearest neighbours of topic-specific embeddings for a sample of words from the WordNet SuperSenses dataset, using BERT-base embeddings. The top 6 selected samples illustrate clear topic distributions per word sense, and the bottom 4 also show topical properties within the same sense. The most relevant words for each topic are shown under the \textbf{\textsc{topic}} column.}
\label{tab:nns}
\end{table*}

\begin{figure*}
\centering
\includegraphics[width=180pt]{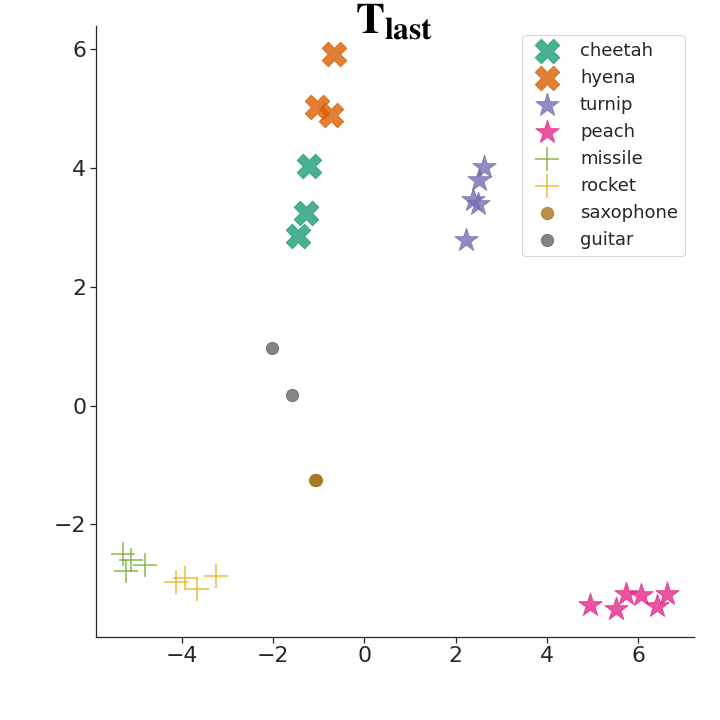}
\hspace{40pt}
\includegraphics[width=180pt]{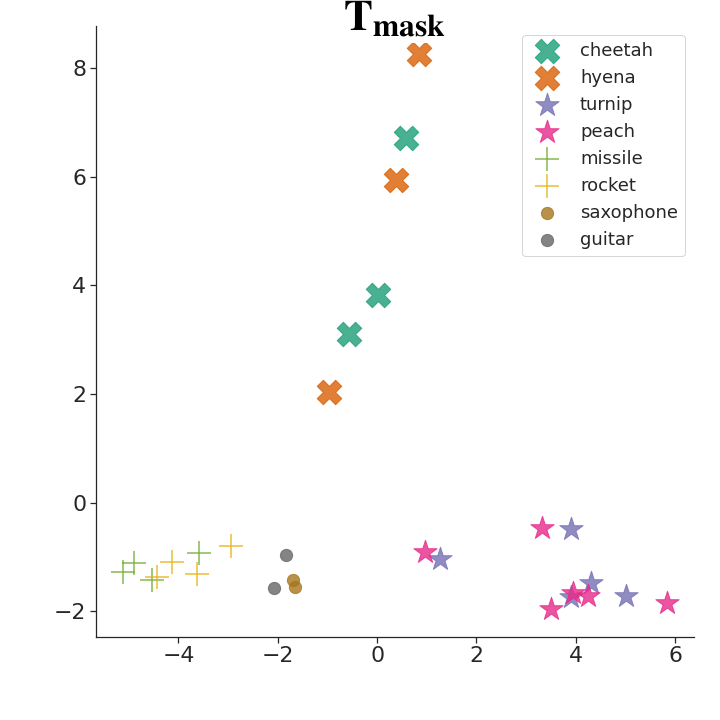}
\caption{BERT-base topic-specific vectors when using the output vectors without using masking (left) and with masking (right). Words have been selected from the McRae dataset.}
\label{figPlot}
\end{figure*}

\subsection{Qualitative analysis}
\label{sec:qualitative}
Topic-specific vectors can be expected to focus on different properties, depending on the chosen topic. In this section, we present a qualitative analysis in support of this view. In Table \ref{tab:nns} we list, for a sample of words from the WordNet supersenses dataset, the top 5 nearest neighbours per topic in terms of cosine similarity. For this analysis, we used the BERT-base masked embeddings. We can see that for the word \emph{`partner'}, its topic-specific embeddings correspond to its usage in the context of `finance', `stock market' and `fiction'. These three embeddings roughly correspond to three different senses of the word\footnote{In fact, we can directly pinpoint these vectors to the following WordNet \cite{miller1995wordnet} senses: {\small{\texttt{partner.n.03}}}, {\small{\texttt{collaborator.n.03}}} and {\small{\texttt{spouse.n.01}}}.}. This de-conflation or implicit disambiguation is also found for words such as \emph{`cell'}, \emph{`port'}, \emph{`bulb'} or \emph{`mail'}, which shows a striking relevance of the role of mail in the election topic, being semantically similar in the corresponding vector space to words such as `telemarketing', `spam' or `wiretap'. In the case of \emph{`fingerprint'}, we can also see some implicit disambiguation (distinguishing between fingerprinting in computer science, as a form of hashing, and the more traditional sense). However, we also see a more topical distinction, revealing differences between the role played by fingerprints in fictional works and forensic research. 
This tendency of capturing different contexts is more evidently shown in the last four examples. First, for \emph{`sky'} and \emph{`strength'}, the topic-wise embeddings do \textit{not represent different senses of these words}, but rather indicate different types of usage (possibly related to cultural or commonsense properties). Specifically, we see that the same sense of \emph{`sky'} is used in mythological, landscaping and geological contexts. Likewise, \emph{`strength'} is clustered into different mentions, but while this word also preserves the same sense, it is clearly used in different contexts: physical, as a human feature, and in military contexts. Finally, \emph{`noon'} and \emph{`galaxy'} (which only occur in two topics), also show this topicality. In both cases, we have representations that reflect their physics and everyday usages, for the same senses of these words.


As a final analysis, In Figure \ref{figPlot} we plot a two-dimensional PCA-reduced visualization of selected words from the McRae dataset, using two versions of the topic-specific vectors: 
\textbf{T\textsubscript{mask}} and \textbf{T\textsubscript{last}}. In both cases, BERT-base was used to obtain the vectors. We select four pairs of concepts which are topically related, which we plot with the same datapoint marker (animals, plants, weapons and musical instruments). 
For \textbf{T\textsubscript{last}}, we can see that the different topic-specific representations of the same word are clustered together, which is in accordance with the findings from \citet{DBLP:conf/emnlp/Ethayarajh19}. For \textbf{T\textsubscript{mask}}, we can see that the representations of words with similar properties (e.g.\ \emph{cheetah} and \emph{hyena}) become more similar, suggesting that \textbf{T\textsubscript{mask}} is more tailored towards modelling the semantic properties of words, perhaps at the expense of a reduced ability to differentiate between closely related words. The case of \emph{turnip} and \emph{peach} is particularly striking, as the vectors are clearly separated in the \textbf{T\textsubscript{last}} plot, while being clustered together in the \textbf{T\textsubscript{mask}} plot.


\section{Conclusions}
We have proposed a strategy for learning static word vectors, in which topic models are used to help select diverse mentions of a given target word and a contextualized language model is subsequently used to infer vector representations from the selected mentions. We found that selecting an equal number of mentions per topic substantially outperforms purely random selection strategies. We also considered the possibility of learning a weighted average of topic-specific vector representations, which in principle should allow us to ``tune'' word representations to different tasks, by learning task-specific topic importance weights. However, in practice we found that a standard average of the topic specific vectors leads to a comparable performance, suggesting that the outperformance of our vectors comes from the fact that they are obtained from a more diverse set of contexts.

\section*{Acknowledgments}
This work was performed using the computational facilities of the Advanced Research Computing @ Cardiff (ARCCA) Division, Cardiff University and using HPC resources from GENCI-IDRIS (Grant 2021-[AD011012273]).


\bibliography{refs}

\begin{thebibliography}{32}
\expandafter\ifx\csname natexlab\endcsname\relax\def\natexlab#1{#1}\fi

\bibitem[{Alghanmi et~al.(2020)Alghanmi, Anke, and
  Schockaert}]{alghanmi2020combining}
Israa Alghanmi, Luis~Espinosa Anke, and Steven Schockaert. 2020.
\newblock Combining bert with static word embeddings for categorizing social
  media.
\newblock In \emph{Proceedings of the Sixth Workshop on Noisy User-generated
  Text}, pages 28--33.

\bibitem[{Amrami and Goldberg(2019)}]{amrami2019towards}
Asaf Amrami and Yoav Goldberg. 2019.
\newblock Towards better substitution-based word sense induction.
\newblock \emph{arXiv:1905.12598}.

\bibitem[{Bianchi et~al.(2020)Bianchi, Terragni, and
  Hovy}]{DBLP:journals/corr/abs-2004-03974}
Federico Bianchi, Silvia Terragni, and Dirk Hovy. 2020.
\newblock Pre-training is a hot topic: Contextualized document embeddings
  improve topic coherence.
\newblock \emph{CoRR}, abs/2004.03974.

\bibitem[{Blei et~al.(2003)Blei, Ng, and Jordan}]{DBLP:journals/jmlr/BleiNJ03}
David~M. Blei, Andrew~Y. Ng, and Michael~I. Jordan. 2003.
\newblock Latent dirichlet allocation.
\newblock \emph{J. Mach. Learn. Res.}, 3:993--1022.

\bibitem[{Bommasani et~al.(2020)Bommasani, Davis, and
  Cardie}]{DBLP:conf/acl/BommasaniDC20}
Rishi Bommasani, Kelly Davis, and Claire Cardie. 2020.
\newblock Interpreting pretrained contextualized representations via reductions
  to static embeddings.
\newblock In \emph{Proceedings ACL}, pages 4758--4781.

\bibitem[{Bordea et~al.(2015)Bordea, Buitelaar, Faralli, and
  Navigli}]{bordea-etal-2015-semeval}
Georgeta Bordea, Paul Buitelaar, Stefano Faralli, and Roberto Navigli. 2015.
\newblock {S}em{E}val-2015 task 17: Taxonomy extraction evaluation
  ({TE}x{E}val).
\newblock In \emph{Proceedings SemEval}, pages 902--910.

\bibitem[{Bordea et~al.(2016)Bordea, Lefever, and
  Buitelaar}]{bordea2016semeval}
Georgeta Bordea, Els Lefever, and Paul Buitelaar. 2016.
\newblock Semeval-2016 task 13: Taxonomy extraction evaluation (texeval-2).
\newblock In \emph{Proceedings SemEval}, pages 1081--1091.

\bibitem[{Camacho-Collados and Navigli(2017)}]{camacho2017babeldomains}
Jose Camacho-Collados and Roberto Navigli. 2017.
\newblock {BabelDomains}: Large-scale domain labeling of lexical resources.
\newblock In \emph{Proceedings EACL}, pages 223--228.

\bibitem[{Chronis and Erk(2020)}]{chronis-erk-2020-bishop}
Gabriella Chronis and Katrin Erk. 2020.
\newblock When is a bishop not like a rook? when it{'}s like a rabbi!
  multi-prototype {BERT} embeddings for estimating semantic relationships.
\newblock In \emph{Proceedings CoNLL}, pages 227--244.

\bibitem[{Devlin et~al.(2019)Devlin, Chang, Lee, and
  Toutanova}]{DBLP:conf/naacl/DevlinCLT19}
Jacob Devlin, Ming{-}Wei Chang, Kenton Lee, and Kristina Toutanova. 2019.
\newblock {BERT:} pre-training of deep bidirectional transformers for language
  understanding.
\newblock In \emph{Proceedings NAACL-HLT}.

\bibitem[{Ethayarajh(2019)}]{DBLP:conf/emnlp/Ethayarajh19}
Kawin Ethayarajh. 2019.
\newblock How contextual are contextualized word representations? comparing the
  geometry of {BERT}, {ELMo}, and {GPT-2} embeddings.
\newblock In \emph{Proceedings EMNLP}, pages 55--65.

\bibitem[{Forbes et~al.(2019)Forbes, Holtzman, and Choi}]{forbes2019neural}
Maxwell Forbes, Ari Holtzman, and Yejin Choi. 2019.
\newblock Do neural language representations learn physical commonsense?
\newblock \emph{Proceedings CogSci}.

\bibitem[{Kolyvakis et~al.(2018)Kolyvakis, Kalousis, and
  Kiritsis}]{kolyvakis2018deepalignment}
Prodromos Kolyvakis, Alexandros Kalousis, and Dimitris Kiritsis. 2018.
\newblock Deepalignment: Unsupervised ontology matching with refined word
  vectors.
\newblock In \emph{Proceedings NAACL-HLT}, pages 787--798.

\bibitem[{Li et~al.(2019)Li, Bouraoui, and Schockaert}]{li2019ontology}
Na~Li, Zied Bouraoui, and Steven Schockaert. 2019.
\newblock Ontology completion using graph convolutional networks.
\newblock In \emph{Proceedings ISWC}, pages 435--452.

\bibitem[{Li et~al.(2016)Li, Chua, Zhu, and Miao}]{DBLP:conf/acl/LiCZM16}
Shaohua Li, Tat{-}Seng Chua, Jun Zhu, and Chunyan Miao. 2016.
\newblock Generative topic embedding: a continuous representation of documents.
\newblock In \emph{Proceedings ACL}.

\bibitem[{Liu et~al.(2020)Liu, McCarthy, and Korhonen}]{liu2020towards}
Qianchu Liu, Diana McCarthy, and Anna Korhonen. 2020.
\newblock Towards better context-aware lexical semantics: Adjusting
  contextualized representations through static anchors.
\newblock In \emph{Proceedings EMNLP}, pages 4066--4075.

\bibitem[{Liu et~al.(2015)Liu, Liu, Chua, and Sun}]{DBLP:conf/aaai/LiuLCS15}
Yang Liu, Zhiyuan Liu, Tat{-}Seng Chua, and Maosong Sun. 2015.
\newblock Topical word embeddings.
\newblock In \emph{Proceedings AAAI}, pages 2418--2424.

\bibitem[{Liu et~al.(2019)Liu, Ott, Goyal, Du, Joshi, Chen, Levy, Lewis,
  Zettlemoyer, and Stoyanov}]{DBLP:journals/corr/abs-1907-11692}
Yinhan Liu, Myle Ott, Naman Goyal, Jingfei Du, Mandar Joshi, Danqi Chen, Omer
  Levy, Mike Lewis, Luke Zettlemoyer, and Veselin Stoyanov. 2019.
\newblock Roberta: {A} robustly optimized {BERT} pretraining approach.
\newblock \emph{CoRR}, abs/1907.11692.

\bibitem[{Ma et~al.(2016)Ma, Cambria, and Gao}]{ma2016label}
Yukun Ma, Erik Cambria, and Sa~Gao. 2016.
\newblock Label embedding for zero-shot fine-grained named entity typing.
\newblock In \emph{Proceedings COLING}, pages 171--180.

\bibitem[{{McRae et al.}(2005)}]{mcrae2005semantic}
Ken {McRae et al.} 2005.
\newblock Semantic feature production norms for a large set of living and
  nonliving things.
\newblock \emph{Behavior research methods}, 37:547--559.

\bibitem[{Mikolov et~al.(2013)Mikolov, Chen, Corrado, and
  Dean}]{DBLP:journals/corr/abs-1301-3781}
Tomas Mikolov, Kai Chen, Greg Corrado, and Jeffrey Dean. 2013.
\newblock Efficient estimation of word representations in vector space.
\newblock In \emph{Proceedings ICLR}.

\bibitem[{Miller(1995)}]{miller1995wordnet}
George~A Miller. 1995.
\newblock Wordnet: a lexical database for {English}.
\newblock \emph{Communications of the ACM}, 38(11):39--41.

\bibitem[{Nikolaev and Kotov(2020)}]{nikolaev2020joint}
Fedor Nikolaev and Alexander Kotov. 2020.
\newblock Joint word and entity embeddings for entity retrieval from a
  knowledge graph.
\newblock In \emph{Proceedings ECIR}, pages 141--155.

\bibitem[{Peinelt et~al.(2020)Peinelt, Nguyen, and
  Liakata}]{DBLP:conf/acl/PeineltNL20}
Nicole Peinelt, Dong Nguyen, and Maria Liakata. 2020.
\newblock {tBERT}: Topic models and {BERT} joining forces for semantic
  similarity detection.
\newblock In \emph{Proceedings ACL}, pages 7047--7055.

\bibitem[{Pennington et~al.(2014)Pennington, Socher, and
  Manning}]{DBLP:conf/emnlp/PenningtonSM14}
Jeffrey Pennington, Richard Socher, and Christopher~D. Manning. 2014.
\newblock {GloVe}: Global vectors for word representation.
\newblock In \emph{Proceedings EMNLP}, pages 1532--1543.

\bibitem[{Reif et~al.(2019)Reif, Yuan, Wattenberg, Vi{\'{e}}gas, Coenen,
  Pearce, and Kim}]{DBLP:conf/nips/ReifYWVCPK19}
Emily Reif, Ann Yuan, Martin Wattenberg, Fernanda~B. Vi{\'{e}}gas, Andy Coenen,
  Adam Pearce, and Been Kim. 2019.
\newblock Visualizing and measuring the geometry of {BERT}.
\newblock In \emph{Proceedings NeurIPS}, pages 8592--8600.

\bibitem[{Reimers and Gurevych(2019)}]{reimers-gurevych-2019-sentence}
Nils Reimers and Iryna Gurevych. 2019.
\newblock Sentence-{BERT}: Sentence embeddings using {S}iamese {BERT}-networks.
\newblock In \emph{Proceedings EMNLP}, pages 3982--3992.

\bibitem[{Shi et~al.(2017)Shi, Lam, Jameel, Schockaert, and
  Lai}]{DBLP:conf/sigir/ShiLJSL17}
Bei Shi, Wai Lam, Shoaib Jameel, Steven Schockaert, and Kwun~Ping Lai. 2017.
\newblock Jointly learning word embeddings and latent topics.
\newblock In \emph{Proceedings SIGIR}, pages 375--384.

\bibitem[{Socher et~al.(2013)Socher, Ganjoo, Manning, and Ng}]{socher2013zero}
Richard Socher, Milind Ganjoo, Christopher~D Manning, and Andrew Ng. 2013.
\newblock Zero-shot learning through cross-modal transfer.
\newblock In \emph{Proceedings NIPS}, pages 935--943.

\bibitem[{Thompson and Mimno(2020)}]{DBLP:journals/corr/abs-2010-12626}
Laure Thompson and David Mimno. 2020.
\newblock Topic modeling with contextualized word representation clusters.
\newblock \emph{CoRR}, abs/2010.12626.

\bibitem[{Vulic et~al.(2020)Vulic, Ponti, Litschko, Glavas, and
  Korhonen}]{DBLP:conf/emnlp/VulicPLGK20}
Ivan Vulic, Edoardo~Maria Ponti, Robert Litschko, Goran Glavas, and Anna
  Korhonen. 2020.
\newblock Probing pretrained language models for lexical semantics.
\newblock In \emph{Proceedings EMNLP}, pages 7222--7240.

\bibitem[{Zhu et~al.(2020)Zhu, Zhou, and He}]{DBLP:journals/tacl/ZhuZH20}
Lixing Zhu, Deyu Zhou, and Yulan He. 2020.
\newblock A neural generative model for joint learning topics and
  topic-specific word embeddings.
\newblock \emph{Trans. Assoc. Comput. Linguistics}, 8:471--485.

\end{thebibliography}
\bibliographystyle{acl_natbib}
\end{document}